# Readability-guided Idiom-aware Sentence Simplification (RISS) for Chinese

**Jingshen Zhang[1], Xinglu Chen[1], Xinying Qiu[1]\*, Zhimin Wang[2], Wenhe Feng[3]**
[1]School of Information Science and Technology,
[2]Faculty of Chinese Language and Culture
[3]Laboratory of Language Engineering and Computing
Guangdong University of Foreign Studies, Guangzhou, China
audbut0702@163.com, luc88749@gmail.com,
xy.qiu@foxmail.com, wangzm000@qq.com, wenhefeng@gdufs.edu.cn

## Abstract

Chinese sentence simplification faces challenges due to the lack of large-scale labeled parallel corpora and the prevalence of idioms. To address these challenges, we propose Readability-guided Idiom-aware Sentence Simplification (RISS), a novel framework that combines data augmentation techniques with lexial simplification. RISS introduces two key components: (1) Readability-guided Paraphrase Selection (RPS), a method for mining high-quality sentence pairs, and (2) Idiom-aware Simplification (IAS), a model that enhances the comprehension and simplification of idiomatic expressions. By integrating RPS and IAS using multi-stage and multi-task learning strategies, RISS outperforms previous state-of-the-art methods on two Chinese sentence simplification datasets. Furthermore, RISS achieves additional improvements when fine-tuned on a small labeled dataset. Our approach demonstrates the potential for more effective and accessible Chinese text simplification.

## 1 Introduction

Sentence simplification (SS) aims to reduce the linguistic complexity of a sentence while preserving its meaning (Al-Thanyyan and Azmi, 2021). It has various societal applications, such as increasing accessibility for people with cognitive disabilities (Stajner, 2021) and second language learners (Petersen and Ostendorf, 2007). While English sentence simplification has made significant progress due to the availability of large-scale parallel corpora, other languages face the challenge of limited publicly available resources (Ryan et al., 2023). For Chinese, in particular, the publicly accessible sentence simplification corpora contain only around 5,000 entries (Sun et al., 2023; Chong et al., 2023). This scarcity of labeled data is a primary hurdle for advancing Chinese sentence simplification research.

In addition to the lack of parallel corpora, Chinese sentence simplification faces another challenge: the prevalence of idioms. Idioms are non-compositional expressions whose metaphorical meanings deviate from the literal meanings of their constituent words (Bobrow and Bell, 1973; Li et al., 2023). In Chinese, idioms are frequently used in sentences, and their simplification often requires providing a brief explanatory sentence rather than simply replacing them with synonyms, as is common in traditional lexical simplification.

To address these challenges, We propose Readability-guided Idiom-aware Sentence Simplification (RISS), a novel framework combining data augmentation and lexical simplification to address the challenges in Chinese sentence simplification. RISS integrates Readability-guided Paraphrase Selection (RPS), which mines high-quality sentence pairs from Chinese paraphrases, and Idiom-aware Simplification (IAS), an idiom-based lexial simplification strategy. Through multi-stage and multi-task learning, RISS outperforms state-of-the-art methods on two datasets without labeled data and further improves when fine-tuned on a small labeled dataset.

---

The paper is organized as follows: Section 2 reviews related work, Section 3 describes our methodologies (RPS, IAS, and RISS), Section 4 outlines the experimental setup, Section 5 presents and analyzes the results, and Section 6 provides conclusions and future directions. The code, data and outputs of the proposed method are available at https://github.com/astro-jon/RISS

## 2 Related Work

### 2.1 Data Augmentation in Sentence Simplification

Data augmentation techniques play a crucial role in addressing the scarcity of training data for sentence simplification, especially for languages other than English. Various approaches have been proposed to create pseudo-parallel corpora, such as mining paraphrases from large-scale datasets (Zhao et al., 2020; Martin et al., 2022), utilizing back-translation (Lu et al., 2021), leveraging summarization datasets (Sun et al., 2023), and using Wikipedia and Simple Wikipedia articles (Kriz et al., 2020). Chi et al.(2023) focused on creating a dataset with sentences at different complexity levels using a paraphrase corpus labeled with difficulty scores.

However, the quality of the augmented data is essential for effectively training simplification models. Factual errors, such as inadequate simplifications or misalignments, can negatively impact model performance (Xu et al., 2015). To mitigate this issue, researchers have proposed methods to identify and down-weight sentence pairs with factual errors during training (Ma et al., 2022), combine rule-based sentence splitting and deletion with paraphrasing (Jiang et al., 2020), and leverage sentence alignment, paraphrasing, and simplification-specific noise reduction techniques (Maddela et al., 2021).

Despite these advancements, there is still a need for methods that can effectively select high-quality paraphrase pairs that are most suitable for the simplification task. Our proposed Readability-guided Paraphrase Selection (RPS) framework addresses this gap by automatically selecting high-quality paraphrase pairs based on readability differences and linguistic similarities.

### 2.2 Lexical Simplification

Lexical simplification is a key component of sentence simplification that aims to replace complex words or phrases with simpler alternatives while preserving the original meaning (Shardlow, 2014). This process is essential for improving the readability and accessibility of texts for various audiences (Paetzold and Specia, 2017). Previous works have explored different approaches to lexical simplification, such as learning explicit editing operations (Dong et al., 2019), neural ranking (Paetzold and Specia, 2017), and leveraging word-complexity lexicons and neural readability ranking models (Maddela and Xu, 2018).

However, lexical simplification faces challenges in handling idioms and non-compositional expressions (Abrahamsson et al., 2014; Rello et al., 2013). Recent research has explored integrating idiomatic knowledge into sentence simplification models (Yimam et al., 2018) and proposed multilingual lexical simplification methods via paraphrase generation (Qiang et al., 2023). Wan et al.(2023) introduced a strategy that dynamically identifies and masks complex words in complicated sentences and simple words in simple sentences to enhance lexical simplification.

Despite these advancements, the simplification of idioms remains a significant challenge due to their figurative and context-dependent meanings (Rello et al., 2013). Our proposed Idiom-aware Simplification (IAS) module addresses this challenge by leveraging idiom paraphrasing to enhance the readability and comprehensibility of simplified text, aligning with the broader goal of making written content more accessible for all readers (Shardlow, 2014).

## 3 Methodology

Our proposed Readability-guided Idiom-aware Sentence Simplification framework combines two novel methodologies: RPS for mining high-quality paraphrase data and IAS for idiom simplification. We integrate these modules using multi-stage and multi-task learning strategies to enhance the performance of sentence simplification models. We provide a glossary of abbreviations in Appendix A.

### 3.1 RPS: Readability-guided Paraphrase Selection

The RPS module, as shown in Figure 1, automatically mines high-quality paraphrase sentence pairs to serve as training data for sentence simplification models. Given a set of paraphrase data, inspired by previous research(Bandel et al., 2022), we calculate pairwise readability differences, Levenshtein similarities, syntactic similarities, and semantic similarities for each sentence pair. These features are then integrated into a quality scoring algorithm that selects the highest-quality(Hi-Q) paraphrase pairs based on predefined thresholds.

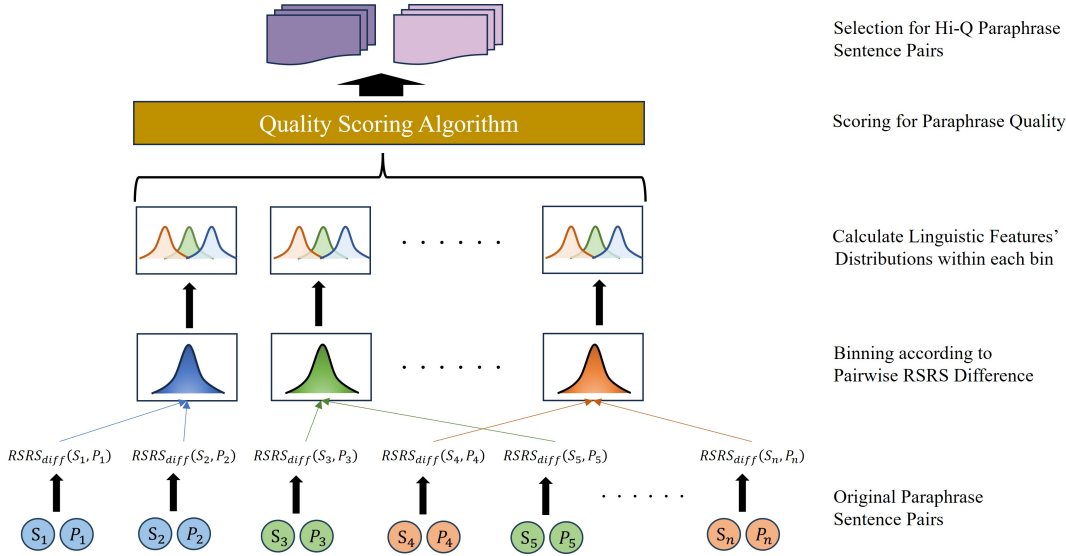

Figure 1: The architecture of the proposed RPS framework

**1. Pairwise Readability Difference:** Given a paraphrase sentence pair $(S_i, P_i)$, we compute their readability scores, $RSRS_i$ and $RSRS_j$, using the $RSRS$ (Martinc et al., 2021), a model evaluation-based readability metric. The RSRS score is calculated as follow:

$$RSRS = \frac{\sum_{i=1}^{S} \sqrt{i} \cdot WNLL(i)}{S} \quad (1)$$

where $S$ represents the length of the sentence, $i$ represents the rank of a word in this sentence (sorted in ascending order by $WNLL$ value), and $WNLL$ is the word-level negative log-likelihood of the $i-th$ word, computed as:

$$WNLL = -(y_t log y_p + (1-y_t) log(1-y_p)) \quad (2)$$

where $y_p$ denotes the probability predicted by the language model based on the historical sequence, and $y_t$ denotes the empirical distribution for a specific position in the sentence, (i.e., $y_t = 1$ for the actual next word and 0 for all the other words in the vocabulary).

We choose distilbert-base-multilingual-cased(Sanh et al., 2019) as the RSRS discriminant model, which has been experimentally proven effective (see Appendix E for details). The pairwise readability difference, defined as $RSRS_{Diff}(S_i, P_i) = RSRS(S_i) - RSRS(P_i)$, quantifies the simplification difference between the original sentence and its paraphrase. Considering this difference allows selecting paraphrase pairs with desirable simplification characteristics to enhance sentence simplification model training.

**2. Pairwise Levenshtein Similarity:** We calculate the Levenshtein similarity between a paraphrase sentence pair $(S_i, P_i)$ as follow:

$$\text{Lev}_{Sim}(S_i, P_i) = \frac{sum(S_i, P_i) - ldist(S_i, P_i)}{sum(S_i, P_i)} \quad (3)$$

where the $sum(S_i, P_i)$ is the total length of $S_i$ and $P_i$, $ldist(S_i, P_i)$ is the weighted edit distance between $S_i$ and $P_i$ based on Levenshtein distance (Levenshtein, 1966). The weighted distance is calculated as :

$$ldist(S_i, P_i) = Num(INSERT) + Num(DELETE) + 2 * Num(REPLACE) \quad (4)$$

where $Num(\cdot)$ denotes the count of the respective edit operation. The motivation behind using Levenshtein similarity is to ensure that the selected paraphrase pairs maintain a sufficient level of lexical similarity, which is important for preserving the core meaning of the original sentence during simplification.

**3. Pairwise Syntactic Similarity:** To measure the syntactic similarity between a paraphrase sentence pair $(S_i, P_i)$, we first construct the third-level constituency parse trees for both sentences. We then calculate the syntactic similarity as follows:

$$\text{Syn}_{Sim}(S_i, P_i) = 1 - \frac{TreeEditDistance(\text{Parse}_{Tree}(S_i), \text{Parse}_{Tree}(P_i))}{\max(\text{len}(\text{Parse}_{Tree}(S_i)), \text{len}(\text{Parse}_{Tree}(P_i)))} \quad (5)$$

where $\text{Parse}_{Tree}(S_i)$ and $\text{Parse}_{Tree}(P_i)$ represent the third-level constituency parse trees of $S_i$ and $P_i$, respectively, and $TreeEditDistance(\cdot, \cdot)$ represents normalized tree edit distance (Zhang and Shasha, 1989). The tree edit distance computes the minimum number of edit operations required to transform one parse tree into the other. Using syntactic similarity helps select paraphrase pairs with similar structures, preserving grammatical integrity in simplified sentences. Pairs with high syntactic similarity provide consistent patterns for learning simplification transformations, making them effective for training sentence simplification models.

**4. Pairwise Semantic Similarity:** To measure the semantic similarity between a paraphrase sentence pair $(S_i, P_i)$, we employ the text2vec[0] embedding approach. We extract the vector representations of both sentences using text2vec model[1] and calculate their cosine similarity as follows:

$$\text{Sem}_{Sim}(S_i, P_i) = \frac{\sum_{j=1}^{n}(text2vec(S_i)_j \times text2vec(P_i)_j)}{\sqrt{\sum_{j=1}^{n} text2vec(S_i)_j^2} \times \sqrt{\sum_{j=1}^{n} text2vec(P_i)_j^2}} \quad (6)$$

where $\text{text2vec}(S_i)$ and $\text{text2vec}(P_i)$ denote the vector representations of $S_i$ and $P_i$, respectively, obtained from the text2vec model, $n$ denotes the dim of embedding.

Using semantic similarity ensures that selected paraphrase pairs preserve the original sentence's core meaning and information during simplification.

**Correlation among features:** Figure 2 shows that higher readability differences correspond to lower linguistic similarity scores in paraphrase pairs. Pearson correlation coefficients confirm strong, statistically significant positive correlations between readability difference and Levenshtein ($r = 0.653$), syntactic ($r = 0.612$), and semantic ($r = 0.624$) similarities ($p < 0.05$). The imperfect correlations ($r < 1$) suggest each feature captures distinct aspects of the original-simplified sentence relationship. These correlations validate our approach of using these features with readability differences to select high-quality paraphrase pairs for sentence simplification task.

---

[0] https://pypi.org/project/text2vec/
[1] https://huggingface.co/shibing624/text2vec-base-chinese

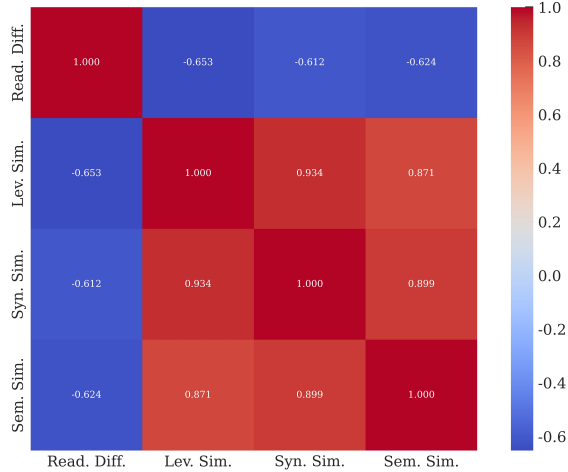

Figure 2: The heat map displays the correlations between readability differences and lexical, syntactic, and semantic similarities. Cool tones indicate that higher readability differences correlate with lower feature values.

**Readability-guided Paraphrase Selection (RPS) Module:** The RPS framework (Figure 1) mines high-quality paraphrase pairs for training sentence simplification models through the following steps:

1. Readability assessment and binning of the paraphrase corpus using the $RSRS_{Diff}$ scores.

2. Extraction of linguistic features ($Lev_{Sim}$, $Syn_{Sim}$, $Sem_{Sim}$) for each pair within each bin.

3. Quality scoring (Eq. 7) to select the best pairs from each bin based on feature means and standard deviations.

$$Q(p,b) = \begin{cases} 1, & \forall_{f \in \{lex, syn, sem\}} (\mu_f^b - \sigma_f^b \leq f(p) \leq \mu_f^b + \sigma_f^b) \\ 0, & \text{otherwise} \end{cases} \quad (7)$$

where $p$ represents a paraphrase sentence pair, $b$ represents a bin, $f$ represents a feature (Lex$_{Sim}$, Syn$_{Sim}$, or Sem$_{Sim}$), $\mu_f^b$ represents the mean of feature $f$ in bin $b$, $\sigma_f^b$ represents the standard deviation of feature $f$ in bin $b$, $f(p)$ represents the scoring value of feature $f$ for pair $p$.

4. Creation of a curated dataset of high-quality pairs for training simplification models. By considering multiple linguistic dimensions and their relationships with readability differences, RPS identifies paraphrase pairs well-suited for simplification while balancing simplification and fidelity to the original meaning.

## 3.2 Idiom-aware Simplificaiton (IAS) Module

As shown Appendix C, we observe a fair amount of idiom prevalance in the Chinsese Sentence Simplification Datasets. This observation motivates us to propose an idom-aware simplification approach. The IAS module (Figure 3) enhances the sentence simplification model's ability to understand and simplify idioms. IAS introduces an idiom-specific loss component to encourage accurate representations and simplifications of idiomatic phrases. During training, IAS augments the standard sentence-level loss (Eq. 8) with an idiomatic loss term (Eq. 9) that focuses on generating correct idiom simplifications. By jointly optimizing both loss terms (Eq. 10)., IAS guides the model to learn general simplification patterns and specific idiom simplification rules, helping it produce more coherent and understandable simplified sentences. Please note that the idiom paraphrase dataset already provides the position of the idiom explanation in the target sentence corresponding to the idiom in the source sentence. Consequently, training an additional model to identify idiom explanations is unnecessary. For more details about locating idioms in calculating idiom-specific loss, please refer to Appendix D.

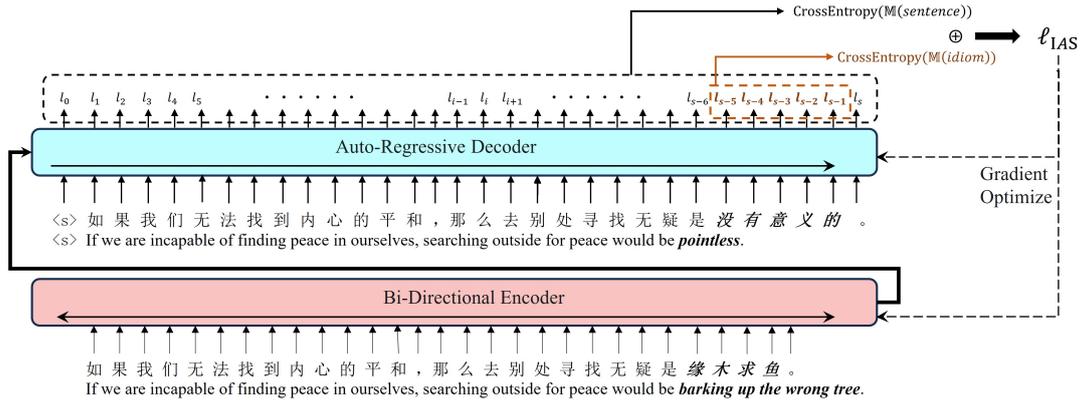

Figure 3: The IAS module architecture. $l_i$ denotes the logits predicted by model $\mathbb{M}$ for the labeled sentence. Bolded italics show the idiom and its explanation.

$$\ell_{\text{sentence}} = \text{CrossEntropy}(\mathbb{M}(\text{source}), \text{target}) \tag{8}$$

where $\mathbb{M}(\cdot)$ denotes the sentence simiplificaton model, source is the input sentencem and target is the ground-truth simplified sentence.

$$\ell_{\text{idiom}} = \text{CrossEntropy}(\mathbb{M}(\text{idiom}), \text{explanation}) \tag{9}$$

where idiom is the idiomatic expression encountered in the source sentence, and explanation is the corresponding simplified version or explanation of the idiom.

$$\ell_{\text{IAS}} = \ell_{\text{sentence}} + \ell_{\text{idiom}} \tag{10}$$

### 3.3 RISS Module

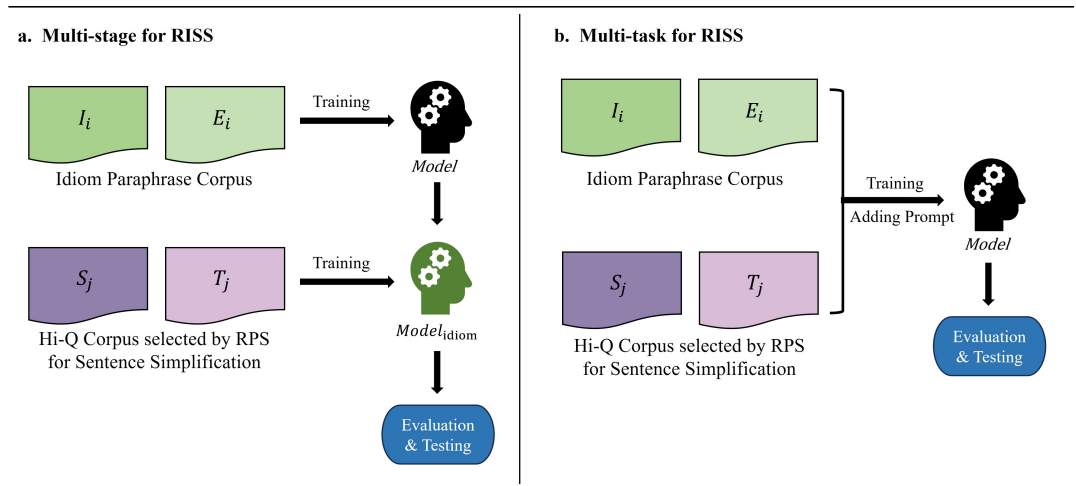

Figure 4: The RISS module architecture with multi-stage (left) and multi-task (right) strategies, where $I_i$, $E_i$ denote idioms and their corresponding explanations respectively, and $S_j$, $T_j$ denote source and target sentences selected by RPS respectively.

Computational LinguisticsThe RISS module (Figure 4) integrates RPS and IAS to address the challenges of limited parallel corpora and idiom prevalence in Chinese sentence simplification. RISS employs two training strategies:

**Multi-Stage Training:**

1. Pre-train the model using the IAS module on the idiom simplification dataset.

2. Fine-tune the pre-trained model using the high-quality paraphrase pairs selected by the RPS module.

**Multi-Task Approach with Prompt:**

1. Prepare the training data by adding task-specific prompts to the input sentences. For the idiom simplification task, the input prompt is: "请解释以下成语的意思："(Please explain the meaning of the following idiom:) For the sentence simplification task, the input prompt is: "请简化以下句子："(Please simplify the following sentence:)

2. Train the model simultaneously on both tasks using the prompted data. Task-specific prompts help the model distinguish between tasks and apply appropriate simplification strategies.

RISS offers a flexible and effective framework for Chinese sentence simplification by combining RPS and IAS through multi-stage or multi-task training.

## 4 Experiment Design

In this section, we present the experimental setup for evaluating the proposed RISS framework. We describe the datasets used for training and evaluation, the evaluation metrics, and the baseline methods compared against RISS. And we train all the models using implementation of Bart-base-chinese (Lewis et al., 2020).

### 4.1 Dataset

We use the following datasets for training and evaluating our Chinese sentence simplification framework:

**Paraphrase for RPS**. We employ the PKU Paraphrase Bank (Zhang et al., 2019) as the source of paraphrase pairs for the RPS module. This dataset contains 509,832 sentence pairs extracted from 95 translations of 40 novels, providing a rich and diverse set of paraphrase examples.

**Idiom for IAS**. For the IAS module, we utilize the Chinese Idiom Paraphrase (CIP) dataset (Qiang et al., 2022). CIP consists of 105,530 idiomatic expressions along with their paraphrases, divided into training (95,530 pairs), validation (5,000 pairs), and testing (5,000 pairs) sets. We use the training set for our idiom simplification experiments.

**Evaluation Datasets:** We evaluate the performance of RISS and the baseline methods on two recently released Chinese sentence simplification datasets:

- **CSS** (Sun et al., 2023) is derived from the PFR Chinese corpus and contains 288 manually simplified sentences for few-shot learning and 766 human simplifications (two per original sentence) for evaluation.

- **MCTS** (Chong et al., 2023) is constructed from the Penn Chinese Treebank (CTB) and includes 366 and 357 original sentences for validation and testing, respectively, each with five human simplifications.

Table 1 provides a statistical summary of the CSS and MCTS datasets.

| Datasets | Train | | Valid | | Test | |
|---|---|---|---|---|---|---|
| | Source | Target | Source | Target | Source | Target |
| CSS | 288 | 288*1 | - | - | 383 | 383*2 |
| MCTS | - | - | 366 | 366*3 | 357 | 357*5 |

Table 1: The statistical analysis of CSS and MCTS. Source denotes original sentences, Target denotes manual simplified sentences and _*x denotes each original sentence has x manually simplified versions.

### 4.2 Evaluation Metrics

To assess the quality of the simplified sentences generated by RISS and the baseline methods, we employ the following automatic evaluation metrics:

**SARI** (Xu et al., 2016) is a commonly used metric for evaluating sentence simplification systems. It calculates the F1 scores for three operations (addition, copying, and deletion) by comparing the system output against the input sentence and multiple human-generated references. Following Wan et al. (2023), we report SARI scores at both character and word levels.

**BERT-Score** (Zhang et al., 2019) measures the semantic similarity between the system output and human references using contextualized embeddings from pre-trained BERT models. We use the bert-base-chinese model from the Hugging Face library to compute BERT-Score.

**BLEU** (Papineni et al., 2002) is a precision-based metric originally designed for machine translation evaluation. It calculates the n-gram overlap between the system output and human references. We use the implementation provided by the NLTK library.

### 4.3 Baselines

We compare the performance of RISS against the following baseline methods:

- **Identity** The original sentences are used as simplification output without any modifications.

- **Truncation** The original sentences are truncated to the first 80% of their words.

- **Direct Back-Translation** (Lu et al., 2021) This method uses back-translation to perform unsupervised sentence simplification, with English as the pivot language.

- **Translate Training** (Chong et al., 2023; Sun et al., 2023) The WikiLarge English sentence simplification dataset (Zhang and Lapata, 2017) is translated into Chinese and used to train a Chinese simplification model.

- **Cross-Lingual Pseudo** (Chong et al., 2023) This method constructs pseudo-parallel data for Chinese sentence simplification by translating a Chinese corpus into English, simplifying the English sentences using an existing model, and then translating the simplified sentences back into Chinese.

- **mT5-large** (Sun et al., 2023)Directly finetune with the additional dataset from CSS.

- **GPT-3.5-turbo** (Feng et al., 2023) A large language model fine-tuned for sentence simplification, applied in a zero-shot or few-shot setting.

## 5 Results and Analysis

In this section, we present the experimental results of the proposed Readability-guided Idiom-aware Sentence Simplification (RISS) framework and compare its performance against the baseline methods on the CSS and MCTS datasets. We also conduct an ablation study to investigate the individual contributions of the RPS and IAS modules.

### 5.1 Main Result

Table 2 presents the main results of our experiments on the CSS and MCTS datasets. We report the best-performing RISS variant (multi-stage or multi-task) for each dataset and compare it with the baseline methods. The main difference between "Unsupervised (zero-shot)" and "Supervised (few-shot)" is whether the model is fine-tuned using the CSS training set. The training data for both settings include the "idiom paraphrase corpus" and "high-quality corpus selected by RPS" (Figure 4). For the MCTS dataset, which lacks training data, we use the CSS training set for fine-tuning in the supervised experiments.

| Method | CSS | | | | MCTS | | | |
|---|---|---|---|---|---|---|---|---|
| | $Sari_{char}$ | $Sari_{word}$ | B.S. | BLEU | $Sari_{char}$ | $Sari_{word}$ | B.S. | BLEU |
| Identity | 29.08 | 27.61 | 0.88 | 88.77 | 25.06 | 22.37 | 0.90 | 84.75 |
| Truncation | 32.95 | 33.18 | 0.79 | 76.36 | 25.21 | 21.85 | 0.87 | 61.6 |
| Gold Reference | 46.72 | 45.71 | 0.96 | 65.31 | 50.12 | 48.11 | 0.90 | 61.62 |
| **Unsupervised (Zero-Shot) Method** | | | | | | | | |
| Lu et al. (2021) | 36.27 | 33.39 | - | 63.47 | - | 40.37 | - | 48.72 |
| Translate Training (2023) | 36.02 | 34.44 | - | 71.41 | - | 28.30 | - | **82.20** |
| Cross-Lingual Pseudo (2023) | - | - | - | - | - | 38.49 | - | 63.06 |
| Gpt-3.5-turbo (2023) | 31.95 | 28.92 | <u>0.83</u> | 42.22 | - | 42.39 | - | 49.22 |
| **RISS**<sub>multi-stage</sub> | <u>40.95</u> | <u>39.36</u> | 0.81 | <u>81.92</u> | <u>44.37</u> | <u>42.71</u> | **0.90** | 79.94 |
| **RISS**<sub>multi-task</sub> | **41.68** | **40.52** | **0.87** | **84.05** | **46.23** | **44.36** | **0.90** | <u>80.83</u> |
| **Supervised (Few-Shot) Method** | | | | | | | | |
| mT5-large (2023) | 37.57 | 35.97 | - | 74.71 | - | - | - | - |
| Gpt-3.5-turbo (2023) | 39.32 | 36.57 | <u>0.85</u> | 60.57 | - | - | - | - |
| **RISS**<sub>multi-stage+labeled data</sub> | **44** | **43.1** | **0.88** | <u>88.48</u> | <u>47.41</u> | <u>45.94</u> | **0.90** | <u>79.94</u> |
| **RISS**<sub>multi-task+labeled data</sub> | <u>43</u> | <u>42.35</u> | **0.88** | **88.59** | **48.49** | **46.42** | **0.90** | **81.12** |

Table 2: Main results of the experiment. For a comprehensive analysis, we calculate the missing metrics for publicly available system-generated results provided by previous studies; otherwise, we use "-" to fill in the missing values. **Bold** indicates the best result, and <u>underline</u> indicates the second-best result. For detailed experiment parameters, please see Appendix B.

**RISS Unsupervised Results**. RISS achieves state-of-the-art performance on both datasets in the unsupervised setting. On CSS, RISS obtains a 6.08 SARI point improvement over the previous best unsupervised method and even outperforms the best supervised method by 2.36 points. On MCTS, RISS surpasses the previous best unsupervised method by 1.97 SARI points. Moreover, RISS maintains high semantic similarity with the original sentences, as evidenced by BERT-Scores above 0.8 on both datasets.

The multi-task variant of RISS consistently outperforms the multi-stage variant in the unsupervised setting. We hypothesize that the multi-task approach allows the model to better leverage knowledge from the large-scale idiom paraphrase dataset while simultaneously learning to perform sentence simplification, enhancing its ability to handle idiomatic expressions.

**RISS Supervised Results**. When incorporating labeled simplification data, RISS further advances the state-of-the-art on both datasets. Although the performance of the multi-stage and multi-task variants is not consistent across CSS and MCTS, both variants achieve significant improvements over the baselines, demonstrating the effectiveness of our proposed framework.

### 5.2 Ablation Study

Table 3 presents an ablation study investigating the individual contributions of the RPS and IAS modules.

**RPS:** In both supervised and unsupervised settings, the RPS module improves overn Raw Paraphrase on both datasets, with the highest character-level SARI score improvement reaching 4.09 points. This highlights the importance of mining high-quality paraphrase pairs.

| Method | CSS | | | | MCTS | | | |
|---|---|---|---|---|---|---|---|---|
| | $Sari_{char}$ | $Sari_{word}$ | B.S. | BLEU | $Sari_{char}$ | $Sari_{word}$ | B.S. | BLEU |
| **Unsupervised (Zero-Shot) Method** | | | | | | | | |
| Raw Paraphrase | 36.45 | 35.22 | 0.806 | 86.68 | 41.27 | 39.8 | **0.908** | 81.1 |
| RPS | 40.54 | 38.98 | 0.867 | 82.51 | 43.89 | 42.29 | 0.90 | 78.24 |
| IAS$_{without\text{-}loss}$ | 36.23 | 35.39 | <u>0.874</u> | <u>89.17</u> | 45.84 | 43.72 | 0.90 | **81.73** |
| IAS$_{with\text{-}loss}$ | 36.56 | 35.89 | **0.878** | **89.66** | **46.31** | **44.76** | <u>0.901</u> | 81.18 |
| RISS$_{multi\text{-}stage}$ | <u>40.95</u> | <u>39.36</u> | 0.811 | 81.92 | 44.37 | 42.71 | 0.899 | 79.94 |
| RISS$_{multi\text{-}task}$ | **41.68** | **40.52** | 0.873 | 84.05 | <u>46.23</u> | <u>44.36</u> | 0.899 | 80.83 |
| **Supervised (Few-Shot) Method** | | | | | | | | |
| Bart | 40.21 | 39.39 | <u>0.886</u> | 88.13 | 44.8 | 43.43 | 0.899 | 81.48 |
| Raw Paraphrase | 41.32 | 40.33 | 0.868 | **89.45** | 45.13 | 44.28 | **0.905** | 81.49 |
| RPS | <u>43.76</u> | <u>42.68</u> | 0.87 | 87.84 | 46.47 | 45.48 | 0.903 | 80.94 |
| IAS$_{without\text{-}loss}$ | 41.51 | 40.73 | 0.877 | 88.34 | 47.37 | 45.71 | 0.903 | **81.68** |
| IAS$_{with\text{-}loss}$ | 42.99 | 42.26 | **0.902** | <u>88.63</u> | 47.62 | 46.11 | 0.895 | 81.15 |
| RISS$_{multi\text{-}stage}$ | **44** | **43.1** | 0.875 | 88.48 | 47.42 | 45.94 | 0.903 | <u>81.57</u> |
| RISS$_{multi\text{-}task}$ | 43 | 42.35 | 0.876 | 88.59 | **48.49** | **46.42** | <u>0.904</u> | 81.12 |

Table 3: Ablation Study on CSS and MCTS. **Bold** indicates the best result, and <u>underline</u> indicates the second-best result.

**IAS:** The IAS module also consistently improves SARI scores, although the gains are smaller compared to RPS. Interestingly, even without the RPS module or labeled simplification data, IAS achieves the best unsupervised SARI score on MCTS, outperforming some supervised methods. We attribute this to the higher proportion of idioms in MCTS (14.3%) compared to CSS (10.4%), suggesting that idiom paraphrasing is particularly effective for datasets with a higher prevalence of idiomatic expressions.

**RISS** The full RISS framework, combining RPS and IAS through either multi-stage or multi-task training, achieves the best or second-best SARI scores in most settings, confirming the effectiveness of our proposed approach.

Previous study has pointed out that the BLEU metric might be misleading in assessing sentence simplification (Sulem et al., 2018), and its shortcomings are particularly evident in Chinese (Sun et al., 2023; Chong et al., 2023). Furthermore, BERTScore primarily measures semantic preservation and dose not effectively indicate the extent of simplification. When evaluating simplification systems, focus should be on SARI scores. We provide additional evaluation metrics to offer more comprehensive analysis. Although IAS achieves higher BERTScore and BLEU scores in the ablation study, the design of the RPS module remains essential for improving overall simplification performance, as evidenced by the SARI scores.

## 6 Conclusion

In this work, we propose Readability-guided Idiom-aware Sentence Simplification (RISS), a novel framework that addresses two major challenges in Chinese sentence simplification: the scarcity of labeled parallel corpora and the prevalence of idioms. RISS combines two key components: Readability-guided Paraphrase Selection (RPS), which mines high-quality paraphrase pairs from a large corpus based on readability differences and linguistic similarities, and Idiom-aware Simplification (IAS), an idiom paraphrasing model that leverages an idiom-specific loss function to enhance the comprehension and simplification of idiomatic expressions. By integrating RPS and IAS through multi-stage and multi-task learning strategies, RISS achieves state-of-the-art performance on two Chinese sentence simplification datasets, even without using labeled simplification data. Furthermore, when fine-tuned on a small labeled dataset, RISS demonstrates further improvements, advancing the state-of-the-art in Chinese sentence simplification. We hope that our work will inspire future research in low-resource sentence simplification and contribute to making written content more accessible to diverse audiences.

## 7 Limitation

While RISS demonstrates state-of-the-art performance in Chinese sentence simplification, there are opportunities for further improvement. The quality scoring algorithm in the RPS module, although effective, could potentially be enhanced to identify and retain an even larger proportion of high-quality paraphrase pairs. Additionally, due to time constraints, we relied solely on automatic evaluation metrics to assess the quality of the system-generated sentences. While these metrics provide valuable insights, a human evaluation would be necessary to comprehensively evaluate the readability, fluency, and meaning preservation of the simplified sentences produced by RISS.

## 8 Acknowledgement

This work is partially supported by Guangzhou Science and Technology Plan Project (202201010729), Major Program of National Social Science Foundation of China (18ZDA295), and Guangdong Social Science Foundation Project (GD24CWY11). We thank the anonymous reviewers for their helpful comments and suggestions.

Computational Linguistics## Appendix A  Glossary of Abbreviations

| | |
|---|---|
| RISS | Readability-guided Idiom-aware Sentence Simplification |
| RPS | Readabilituy-guided Paraphrase Selection |
| IAS | Idiom-aware Simplification |
| RISS_multi-stage | A multi-stage training strategy where the model is first pre-trained using the idiom paraphrase dataset (IAS module) and then trained on the mined paraphrase dataset (RPS module). |
| RISS_multi-task | A multi-task training strategy where the idiom paraphrase dataset and the mined paraphrase dataset are used together for training. Each sentence is prefixed with a corresponding prompt. |
| RISS_multi-stage+labeled data | A multi-stage training strategy similar to RISS_multi-stage, but with an additional fine-tuning step using the standard labeled dataset after saving the best model from the pre-training stage. |
| RISS_multi-task+labeled data | The best model from the multi-task learning strategy (RISS_multi-task) is saved and then fine-tuned using the standard labeled dataset. |
| Raw Paraphrase | The publicly available large-scale paraphrase corpus is directly used for pre-training without any quality filtering. |
| RPS (as in Table 3) | The publicly available large-scale paraphrase corpus is quality-filtered using the proposed RPS module, and the remaining high-quality paraphrase data is used for pre-training. |
| IAS_without-loss | The model is pre-trained using the publicly available idiom paraphrase corpus without introducing the idiom-specific loss. |
| IAS_with-loss | The model is pre-trained using the publicly available idiom paraphrase corpus while introducing the idiom-specific loss to further enhance the model's ability to understand and handle idioms. |

## Appendix B  Implementation Details

In our work, we train all models using PyTorch and the implementation of Bart-base-chinese from Hugging Face. All models are trained on an NVIDIA GeForce RTX 3080, using the Trainer provided by the Transformers library as our experimental environment. Table 4 presents our specific experimental hyper-parameters. Following Sun et al.(2023), we filtered out sentences with less than 30 tokens when processing the paraphrase corpus. Additionally, we removed sentence pairs with an RSRS difference lower than 0.1, an empirical value, to prevent inadequate simplification.

| Hyper-Parameter | Value | | |
| --- | --- | --- | --- |
| | Multi-stage Pretrain | Multi-task Pretrain | SS Finetune |
| Max source length | 256 | 256 | 256 |
| Max target length | 256 | 256 | 256 |
| Batch size | 4 | 4 | 4 |
| Learning rate | 2e-5 | 2e-5 | 2e-5 |
| Grad. accu. steps | 2 | 2 | 2 |
| Evaluation strategy | steps | steps | steps |
| Eval step | 500 | 1000 | - |
| Optim metric | Sari | Sari | Sari |
| Max steps/epoch | 50000 | 100000 | 25 |

Table 4: Detailed hyper-parameter settings for multi-stage (i.e., using idiom paraphrase or the PKU paraphrase for pretraining respectively.), multi-task pretrain, and the few-shot finetune in our work. For the other hyper-parameters, we used the default values provided by the Transformers library.

## Appendix C  Statistical Analysis of idioms in CSS and MCTS

| Dataset | Train | Val | Test |
| --- | --- | --- | --- |
| CSS | 0.347% | 10.44% | - |
| MCTS | - | 14.48% | 14.28% |

Table 5: Statistical analysis of idioms in CSS and MCTS, where "-" indicates that the corresponding data split is not available in the dataset, making it impossible to calculate the idiom statistics.

## Appendix D  Locating idiom position for calcualting idiom-specific loss

| Source | 如果我们无法找到内心的平和，那么去别处寻找无疑是缘木求鱼。[SEP] 如果我们无法找到内心的平和，那么去别处寻找无疑是 <extra_id_0>。 |
| --- | --- |
| Target | <extra_id_0> 没有意义的 [null]<extra_id_1> |

Table 6: A specific example from CIP, where "<extra_id_0>" in Source denotes the idiom "缘木求鱼" in the original sentence, and followed by "<extra_id_0>" in Target is the corresponding explanation "没有意义的". "[null]" in this context does not carry any particular meaning.

In the source sentence, <extra_id_0> represents the idiom "缘木求鱼", while in the target sentence, the explanation "没有意义的" immediately follows <extra_id_0>.

When calcuating the idiom-specific loss, we do not need to train an additional model to identify idiom explanations. Instead, we can obtain the expected idiom explanation and its corresponding index by writing simple splitting, matching, and replacement code. We also construct a dictionary with idioms as keys and the corresponding explanation indices as values for the additional idiom loss during training.

The detailed process is as follows (using the example in Table 6):

1. Obtain the target sentence and the corresponding idiom explanation:
Extract the sentence containing the idiom mask from the source using [SEP] as the separator, and extract the corresponding explanation from the target. Replace "<extra_id_0>" with the explanation to obtain the target sentence with the idiom explanation.

2. Obtain the indices of the idiom explanation in the target sentence:
Encode the target sentence using a tokenizer to get a series of indices in the PLM vocabulary. Match the indices of the idiom explanation in the target sentence to determine its position.

3. During training, precisely locate the idiom explanation position using the dictionary and calculate the cross-entropy loss.

In summary, our approach leverages the provided idiom paraphrase dataset to accurately identify and simplify idioms without the need for additional training. The idiom-specific loss is calculated using the indices of the idiom explanation in the target sentence, ensuring the model's ability to handle idioms effectively.

## Appendix E  Unsupervised RSRS Metric

The performance of $RSRS$ depends on the evaluation model. To determine the best model, we downloaded pre-trained language models (PLMs) with different architectures from Hugging Face[2] and performed the complexity binary prediction problem according to the $RSRS$ values on CSS dataset. Based on the prediction results, we chose distilbert-base-multilingual-cased as our $RSRS$ discriminant model. Table 7 presents detailed results.

Additionally, we calculated the mean RSRS value for source and reference sentences in CSS dataset, which were 20.00 and 17.99, respectively. A pairwise t-test revealed a significant difference in RSRS between the original and reference sentences ($p < 0.05$), further validating the effectiveness of RSRS.

| PLMs | Accuracy | F1-score |
| --- | --- | --- |
| bert-base-multilingual-uncased | 0.75 | 0.857 |
| bert-base-chinese | 0.51 | 0.676 |
| chinese-roberta-wwm-ext | 0.573 | 0.728 |
| xlm-roberta-base | 0.66 | 0.795 |
| xlm-roberta-base-language-detection | 0.549 | 0.709 |
| **distilbert-base-multilingual-cased** | **0.785** | **0.879** |

Table 7: We evaluated six different PLMs for RSRS on the CSS dataset. The experimental results show that distilbert-base-multilingual-cased achieved an F1-score of 0.879 and an accuracy of 0.785, outperforming the other five models.

---

[2] https://huggingface.co/

## Appendix F  Statistical Analysis of Our RPS Data and CSS

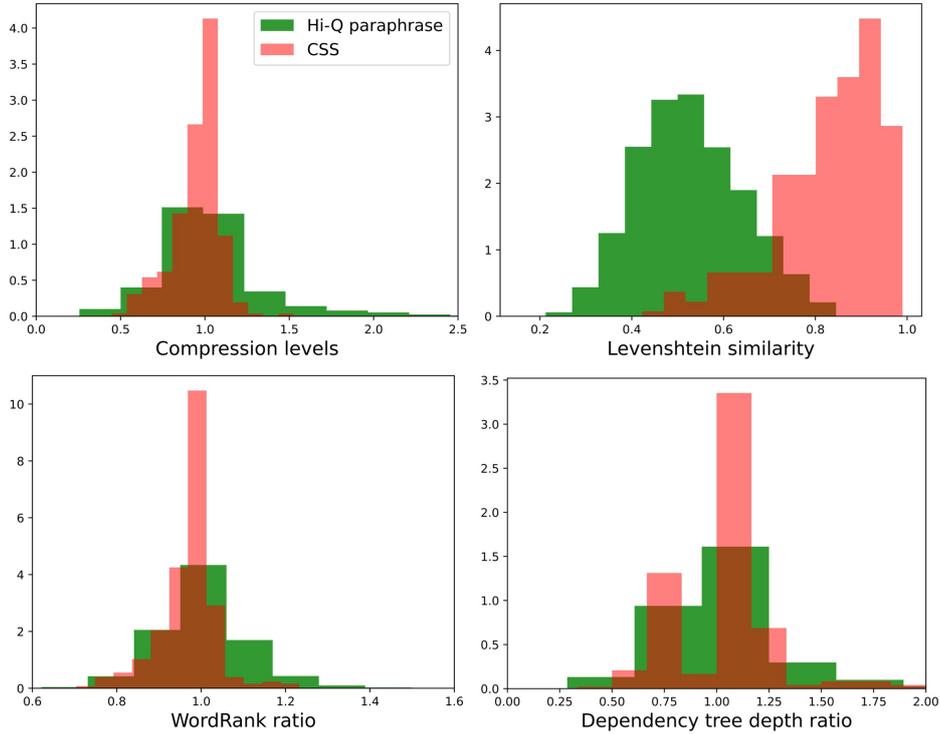

Figure 5: Sentence feature density in the CSS dataset and the data mined by our RPS module. The WordRank ratio is a measure of lexical complexity reduction.

## Appendix G  SARI Evaluation Details

---

**Algorithm 1:** Calculation Process for SARI metric

**Input**    : Original sentences List $orig\_sents$, System generated sentences List $sys\_sents$, Number of Reference version $x$, Reference sentences List(s) $refs\_sents_i$, $1 \leq i \leq x$.

**Output:** Sari score on CSS $Sari_{css}$, Sari score on MCTS $Sari_{mcts}$.

1  $Sari\_list \leftarrow [\,]$;
2  **for** $idx, orig\ in\ enumerate(orig\_sents)$ **do**
3  $\quad orig, sys \leftarrow [orig], [sys\_sents[idx]]$;
4  $\quad refs \leftarrow [[refs\_sents_1[idx]], [refs\_sents_2[idx]], ..., [refs\_sents_x[idx]]]$;
5  $\quad sari \leftarrow$ compute sari $(orig, sys, refs)$;
6  $\quad Sari\_list$.append$(sari)$;
7  **end**
8  $Sari_{css} \leftarrow mean(Sari\_list)$;
9  $Sari_{mcts} \leftarrow$ compute sari $(orig\_sents, sys\_sents, [refs\_sents_1, refs\_sents_2, ..., refs\_sents_x])$;
10 return $Sari_{css}, Sari_{mcts}$

---

Figure 6: Pseudo code for computing sari metric on CSS and MCTS.

Note that document- and sentence-level inputs can lead to different results even if the model generates the same results. Please refer to the pseudo code in Figure 6.

## Appendix H  Case Study

| | |
|---|---|
| Original | 朱老太太原本是上海人，有着上海人的优越感，做事井井有条，说话轻声细语。<br>Mrs. Zhu was originally from Shanghai and had the superiority of Shanghai natives, doing things in a well-organized manner, and speaking with a gentle voice. |
| Gpt-3.5-turbo | 朱老太太原本是上海人，做事**有条不紊**，说话轻声细语，**有**上海人的优越感。<br>Mrs.Zhu was originally from Shanghai, methodically in her work and spoke with a gentle voice, having the superiority of Shanghai natives. |
| $RISS_{multi-stage}$ | 朱老太太**原是**上海人，有着**上海**的优越感，做事**很有条理**，说话轻声细语。<br>Mrs.Zhu was originally from Shanghai and had the superiority of Shanghai natives, doing things very methodically, and speaking with a gentle voice. |
| $RISS_{multi-task}$ | 朱老太太原本是上海人，有着**上海**的优越感，做事**很有条理**，说话**很温和**。<br>Mrs.Zhu was originally from Shanghai and had the superiority of Shanghai natives, doing things very methodically, and speaking gently. |
| $RISS_{multi-stage+labeled\ data}$ | 朱老太太原本是上海人，有着**上海一般人**的优越感，做事**十分有条理**，说话轻声细语。<br>Mrs.Zhu was originally from Shanghai and had the superiority of Shanghai natives, doing things very methodically, and speaking with a gentle voice. |
| $RISS_{multi-task+labeled\ data}$ | 朱老太太原本是上海人。**她**有着**上海**的优越感，做事**很有条理**，说话**很轻声**。<br>Mrs.Zhu was originally from Shanghai. She has the superiority of Shanghai natives, doing things very methodically, and speaking gently. |
| $Reference_1$ | 朱老太太原本是上海人，做事**很有条理**，说话轻声细语。<br>Mrs. Zhu was originally from Shanghai, doing things very methodically, and speaking with a gentle voice. |
| $Reference_2$ | 朱老太太**是**上海人，有着上海人的优越感。**她**做事井井有条，说话轻声细语。<br>Mrs. Zhu was from Shanghai and had the superiority of Shanghai natives. She did things in a well-organized manner, and speaking with a gentle voice. |

Table 8: Generated example from the CSS, where the red font denotes the part of the original sentence that has been simplified by system.

As shown in Table 8, both unsupervised and supervised RISS can identify and simplify the idiom "井井有条"(in a well-organized manner) to "有条理" (methodically). Furthermore, the multitask RISS can rephrase "轻声细语" (in a gentle voice) as "温和" (gently). After incorporating the labeled dataset, RISS further demonstrates clause awareness in its simplifications.